# EVALUATION DE TECHNIQUES DE TRAITEMENT DES REFUSÉS POUR L'OCTROI DE CRÉDIT

Emmanuel Viennet *, Françoise Fogelman Soulié ** & Benoît Rognier **

* Université Paris 13, Institut Galilée, LIPN UMR 7030 CNRS
99, Avenue J-B. Clément - 93430   Villetaneuse
Emmanuel.Viennet@lipn.univ-paris13.fr, http://www-lipn.univ-paris13.fr
** Kxen, 25 Quai Gallieni, 92 158 Suresnes cedex
Francoise.SoulieFogelman@kxen.com, Benoit.Rognier@kxen.com, http://www.kxen.com

**Résumé.** Nous présentons la problématique du traitement des refusés dans le cadre de l'octroi de crédit. Du fait de la législation (Bâle II), les organismes de crédit doivent mettre en place des techniques systématiques d'octroi de crédit et de traitement des refusés. Nous présentons ici une méthodologie de comparaison de différentes techniques de traitement des refusés et montrons qu'il est nécessaire, en l'absence de résultat théorique solide, de pouvoir produire et comparer des modèles adaptés aux données (sélection du "meilleur" modèle conditionnellement aux données). Nous décrivons quelques simulations sur une base de petite taille pour illustrer la démarche et comparer différentes stratégies de choix du groupe de contrôle, qui reste la seule technique fondée de traitement des refusés.

**Abstract.** We present the problem of « Reject Inference » for credit acceptance. Because of the current legal framework (Basel II), credit institutions need to industrialize their processes for credit acceptance, including Reject Inference. We present here a methodology to compare various techniques of Reject Inference and show that it is necessary, in the absence of real theoretical results, to be able to produce and compare models adapted to available data (selection of "best" model conditionnaly on data). We describe some simulations run on a small data set to illustrate the approach and some strategies for choosing the control group, which is the only valid approach to Reject Inference.
**Mots-clés.** Data mining, choix de modèles, risque crédit.

## Introduction

Le traitement des refusés (ou *Reject Inference*) est un problème rencontré de façon systématique dans les activités bancaires d'octroi de crédit. Ce problème est lié au fait que les scores d'octroi sont construits sur les historiques de données disponibles (clients de la banque ayant obtenu et remboursé un crédit, en faisant ou non défaut sur une ou plusieurs mensualités) ce qui fait que l'échantillon des données utilisées pour le calibrage du modèle est systématiquement biaisé en recrutement puisque seuls les Acceptés en font partie. Bien qu'on trouve dans la littérature de très nombreuses techniques de traitement des Refusés, il n'en existe aucune réellement fondée théoriquement, sinon à choisir un groupe de contrôle. On est donc amené, dans la pratique, soit à en choisir une, plus ou moins arbitrairement, soit à comparer, pour chaque nouvelle situation (crédit sur un certain produit), un ensemble de techniques dont la meilleure (conditionnellement à ces données) sera dorénavant utilisée pour ce score produit. La difficulté réside alors dans la capacité à produire facilement de nombreux modèles et à les comparer efficacement.

Nous présentons ici une méthodologie systématique de traitement des refusés, basée sur la production de modèles selon différentes techniques classiques et leur évaluation. Nous utilisons le logiciel KXEN qui permet, très rapidement, de produire les modèles, quelle que soit la volumétrie des données. Nous présentons ici quelques résultats de simulation sur une petite base de données et montrons qu'il est préférable d'utiliser des indicateurs locaux (taux de défaut parmi les acceptés, par exemple) qui seuls permettent d'évaluer le risque du score d'octroi (Hand, 2005).

## Le problème

La plupart des grands organismes de crédit (crédit immobilier ou crédit à la consommation) utilisent des techniques de scoring pour décider de l'opportunité d'accorder leurs prêts et évaluer leurs risques. Les techniques de *score d'octroi* permettent de prendre la décision d'accepter ou refuser le prêt et d'estimer le risque d'un dossier au moment où celui-ci est déposé.

Nous utiliserons les notations suivantes :
- Un dossier de demande de prêt est caractérisé par un vecteur de *k caractéristiques*
    $x = (x_1, x_2, ..., x_k)$
- Le *résultat* de l'emprunt est noté $y \in \{0,1\}$ : $y = 0$ si le prêt finit en défaut, $y = 1$ sinon
- *d* est la *décision* d'octroi : $d \in \{0,1\}$, $d = 1$ si le prêt est accordé, $d = 0$ sinon

Evidemment, *y* n'est connu que pour les clients à qui on a accordé un crédit dans le passé, c'est-à-dire ceux pour lesquels *d=1*. Pour les clients refusés, l'information du résultat est *manquante* : l'ensemble de données disponible a donc un *biais de recrutement* intrinsèque.

Les données clients (Hand, 2001) sont utilisées pour construire un score, *S(x)*, c'est-à-dire une estimation de la probabilité de non-défaut : $S(x) = \hat{P}(y=1/x)$.

La décision d'octroi est alors basée sur le score *S* et sur un seuil *s* :
    $d = 1$ si : $S(x) \geq s$ et $d = 0$ sinon.

En faisant varier *s*, et pour un même score *S*, on pourra faire varier le taux d'acceptation.

Nous nous plaçons donc ici dans le cadre suivant : étant donné un ensemble de clients Acceptés $A = \{i = 1, ..., N\}$ et Refusés $R = \{i = N+1, ..., N+M\}$, caractérisés par un vecteur $\{x^i = (x^i_1, x^i_2, ..., x^i_k), y^i\}$ pour les Acceptés et $\{x^i = (x^i_1, x^i_2, ..., x^i_k)\}$ pour les refusés, le score *S* est calculé en utilisant les données des Acceptés : $\{[x^i = (x^i_1, x^i_2, ..., x^i_k), y^i], i = 1, ..., N\} \rightarrow S$

Le score est ensuite appliqué à tous les dossiers de demande de crédit, qui seront Acceptés ou Refusés selon que $S(x)$ est supérieur ou inférieur au seuil fixé *s*. Une erreur systématique est donc commise, *sample selection error* ou *censoring*, qui peut sous- ou sur-estimer le risque de défaut.

Le *traitement des refusés* ou *Reject Inference* a pour but de corriger ce biais.

## Techniques de traitement des refusés – état de l'art

### Contexte

Il existe de très nombreuses techniques dans la littérature pour construire un score, mais toutes semblent produire des résultats comparables (Baesens et al. 2003, Thomas et al. 2005). Dans les expériences décrites plus loin, nous utiliserons une régression régularisée produite par le module K2R de KXEN. De même, les techniques de traitement des refusés sont très nombreuses (Ash et Meester, 2002, Crook et Banasik 2004, Feelders, 2000, 2003, Hand, 2001) : cependant, la plupart des résultats publiés semblent indiquer des résultats pour le moins mitigés, l'efficacité de la méthode semblant dépendre fortement de la nature des données utilisées. La meilleure méthodologie de traitement des refusés est donc la mise en oeuvre systématique d'un ensemble de méthodes, leur comparaison et la sélection de la méthode la plus adaptée à l'ensemble de données particulier disponible. En effet, en pratique, la banque dispose d'un ensemble de données fixe et la question pour elle n'est pas de déterminer la meilleure méthode en général (qui, aujourd'hui, n'existe pas), mais bien celle qui est la meilleure pour cet ensemble de données là. Ces deux problèmes (inconditionnel et conditionnel, selon la terminologie discutée par Hand, 2005) sont différents et peuvent tout à fait donner des réponses opposées. La seule méthode correcte de traitement des refusés serait bien sûr d'obtenir l'information manquante (Hand, 2001) : en octroyant un crédit à un groupe de contrôle représentatif de la population globale, quelle que soit leur note de score, on peut construire un modèle applicable à l'ensemble de la population. Bien que cette méthode ait un coût, ce coût peut sans doute être compensé par une amélioration de l'estimation du risque par le modèle obtenu (Viennet et Fogelman, 2006).

Le *processus de traitement des refusés* est présenté figure 1 : une première méthode de scoring a abouti à la génération d'une population d'Acceptés (parmi lesquels certains se sont trouvés en défaut par la suite, et d'autres pas) et de Refusés. Il s'agit d'améliorer cette technique en utilisant les refusés, de façon à corriger le biais d'échantillon : les nouveaux dossiers seront ensuite scorés avec ce nouveau modèle. Pour cela, on simule un sous-ensemble de refusés dans l'ensemble des Acceptés, et on calibre plusieurs modèles qu'on compare de façon à choisir le plus adapté, qui est adapté en fonction des refusés initiaux puis appliqué aux nouveaux dossiers.

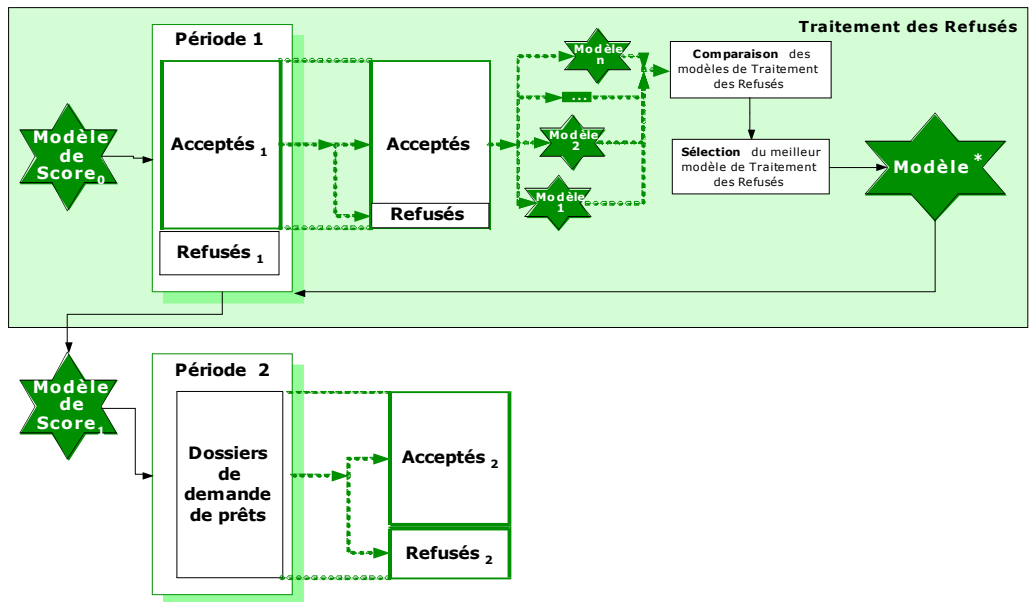

**Figure 1 – Processus de traitement des Refusés**

**Techniques classiques de traitement des refusés**

Les techniques présentées ci-dessous sont les techniques utilisées classiquement dans les institutions de crédit; aucune n'est valide théoriquement, sauf la dernière, à condition que le groupe de contrôle choisi soit représentatif de la population. On distingue dans ce qui suit le score de défaut qui vise à estimer $P(y=1/x)$ et le score d'acceptation qui estime $P(d=1/x)$.

1. *Extrapolation* : on construit un modèle de score de défaut sur les Acceptés et on l'applique à l'ensemble de la population, en extrapolant donc sur la population des refusés.

2. *Reclassification* (« augmented data set ») : on construit un modèle de score de défaut sur les Acceptés et on l'applique aux Refusés (qui reçoivent donc une étiquette défaut ou non-défaut). On construit alors un nouveau score sur la population globale – dite ensemble augmenté – Acceptés + Refusés étiquetés. Cette technique fait implicitement l'hypothèse que $P(y/d=1) = P(y/d=0)$.

3. *Augmentation* (« re-weighting ») : on construit d'abord un score d'acceptation sur l'ensemble des Acceptés + Refusés. Ensuite, on pondère, dans chaque bande de score, chaque Accepté par un poids inverse de la fréquence d'Acceptés dans la bande. Un score de défaut est ensuite construit sur les Acceptés ainsi pondérés.

4. *Parcelling* : on construit un score de défaut sur les Acceptés. On calcule, dans chaque bande de score, le nombre de défaut / non-défaut dans la population des Acceptés. On fait ensuite une hypothèse de taux de défaut des refusés par bande de score et on en déduit le nombre de défaut / non-défaut dans les Refusés de chaque bande. On étiquette ensuite au hasard les Refusés de chaque bande en respectant les proportions. On constitue ainsi un ensemble augmenté et on construit un nouveau score sur la population globale Acceptés + Refusés, comme en 2.

5. *Groupe de contrôle* : on accepte tous les dossiers d'un groupe de contrôle constitué de façon à représenter la population complète (Thomas el al. 2005). On construit ensuite un modèle de score sur cet échantillon. Quand on accepte des Refusés, c'est à dire des dossiers à fort potentiel de défaut, on encourt évidemment un risque accru : les organismes de crédit recrutent donc souvent le groupe de contrôle en essayant de limiter ce risque. Néammoins, cette technique est la meilleure – et la seule – technique de Traitement des Refusés (Hand, 2005), et, en pratique, on contrôle ce risque en regard du bénéfice attendu.

## Expériences

Nous avons mené quelques expériences pour illustrer ce que pourrait être une méthodologie systématique de traitement des refusés. Le but de ces expériences n'est pas d'évaluer la performance finale retenue ou de démontrer la supériorité d'une technique sur une autre, mais bien de démontrer, qu'en l'absence de technique statistiquement fondée (Hand & Henley, 1993), la seule méthode réellement efficace reste de produire des modèles et de déterminer la meilleure solution conditionnée aux données disponibles (Hand, 2005).

Nous utilisons pour nos expériences un jeu de données réelles fourni par un client – anonyme – de KXEN. Ces données concernent l'évaluation du risque crédit et comprennent une trentaine de variables pour environ 6 000 clients. Cet ensemble de données réelles est de très petite taille, tant par le nombre réduit de variables que par le faible nombre de clients. Les résultats obtenus ne sont donc pas très significatifs, et la robustesse des modèles est faible (ce qu'indique le critère de robustesse *KR*, écart entre les courbes de lift des ensembles d'estimation et de validation, calculé automatiquement par KXEN. Dans les expériences décrites plus bas *KR* est de l'ordre de 0,90%). Nous simulons la population des Refusés en choisissant le segment à plus fort taux de défaut d'une segmentation grossière et en y ajoutant 10% de clients tirés au hasard. A l'issue de ce processus, artificiel mais vraisemblable dans le cadre de cette simulation, nous avons 5% de clients refusés. Afin d'estimer la performance et la robustesse des modèles, nous séparons la base en deux groupes : estimation (calibrage des modèles) et validation (comparaison). Nous avons ensuite mis en oeuvre les techniques précédentes à l'aide des modules d'encodage automatique K2C et de classification / régression K2R du logiciel KXEN. Ces modules s'appuient sur les techniques de SRM (Structural Risk Minimization) de Vapnik (Vapnik, 1995), ce qui garantit un compromis optimal entre précision et robustesse des modèles calculés. Pour chacun des modèles présentés ici, les temps de calcul ont été de quelques heures (manipulation de données) et quelques minutes (production du modèle).

Nous avons montré (Viennet et Fogelman, 2006) que, pour comparer les performances des différents modèles, les indicateurs globaux (AUC, KS, GINI, KI, taux de classification) n'étaient pas adaptés : en effet, en octroi de crédit, on travaille toujours autour d'un point de fonctionnement (seuil, taux d'acceptation) et c'est dans le voisinage de ce point qu'il convient de comparer les modèles : on peut ainsi utiliser le *taux de défaut sur les Acceptés* (Hand 2005) ou le revenu global attendu.

La comparaison des techniques précédentes pour une décomposition des 6 000 clients en 40% en estimation et 20% en validation montre que ces techniques sont pratiquement équivalentes ... à ne rien faire : c'est à dire extrapoler (fig. 2 : les barres d'erreur, calculées sous une hypothèse de loi binômiale, sont similaires pour tous les modèles et représentées uniquement pour le modèle M5 de groupe de contrôle).

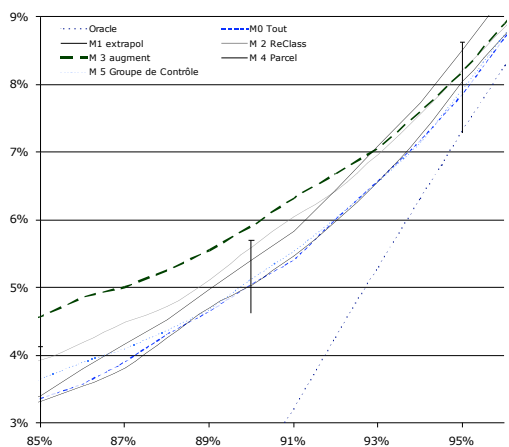

**Figure 2 – Comparaison des taux de défaut en fonction du taux d'acceptation**

Nous avons également testé notre processus global de traitement des refusés (fig. 1) : nous découpons l'ensemble disponible de 6 000 clients en :

- Une partie (A1 : 70%) que nous utilisons pour construire et comparer des modèles. Nous utilisons cinq techniques de traitement des refusés : l'extrapolation (M1), l'augmentation (modèle M2) et 3 méthodes de choix du groupe de contrôle :
  + M3 GC1: on tire au hasard 30% des Refusés de A1;
  + M4 GC2 : on choisit 30% des Refusés de A1 en favorisant les meilleurs dossiers. On applique le score calculé sur A1 aux Refusés, en construisant 20 bandes de score, qu'on numérote de 1 à 20, du plus haut au plus faible score. On tire ensuite, dans la bande *n* (*n=1* à *20*), un nombre d'individus proportionnel à 1/n. On inclue donc davantage de bons dossiers de Refusés : c'est la stratégie souvent utilisée pour limiter le risque et donc le coût du groupe de contrôle.
  + M5 GC3 : comme GC2, mais en favorisant les mauvais dossiers (il suffit de numéroter les bandes de score en ordre ascendant du score).
  
  Le groupe GC2 est évidemment moins risqué que le groupe GC3, qui apporte surtout de l'informa-tion sur les très mauvais dossiers. Ni GC2, ni GC3 ne sont représentatifs de la population.
  
  La comparaison des modèles sur A1 permet de choisir le meilleur modèle.
- Une partie (A2 : 30%) que nous utiliserons pour évaluer le modèle choisi (simulant la "période 2" de la figure 1).

La comparaison de ces techniques montre encore qu'elles sont pratiquement équivalentes que ce soit pour les indicateurs globaux (fig. 3) ou pour le taux de défaut (fig. 4, gauche), ce qui n'est pas surprenant, vu la faible taille des échantillons (dans une application réelle, les bases seraient au moins 10 fois plus grandes).

| Model | AUC | | KS | | GINI | | KI | | ClassificationRate | | KR |
|---|---|---|---|---|---|---|---|---|---|---|---|
| | *Estim.* | *Valid.* | *Estim.* | *Valid.* | *Estim.* | *Valid.* | *Estim.* | *Valid.* | *Estim.* | *Valid.* | |
| M1 A1 | 0,899 | 0,848 | 0,661 | 0,578 | 0,738 | 0,645 | 0,798 | 0,696 | 0,927 | 0,930 | 0,899 |
| M2 augment | 0,904 | 0,859 | 0,669 | 0,583 | 0,737 | 0,658 | 0,807 | 0,719 | 0,924 | 0,918 | 0,912 |
| M3 GC1 | 0,908 | 0,847 | 0,694 | 0,578 | 0,743 | 0,643 | 0,817 | 0,694 | 0,920 | 0,929 | 0,877 |
| M4 GC2 | 0,899 | 0,849 | 0,684 | 0,578 | 0,727 | 0,647 | 0,797 | 0,698 | 0,921 | 0,927 | 0,901 |
| M5 GC3 | 0,907 | 0,844 | 0,689 | 0,578 | 0,739 | 0,638 | 0,814 | 0,689 | 0,919 | 0,924 | 0,875 |

**Figure 3 – Comparaison des indicateurs globaux sur A1 (ensembles d'estimation et de validation)**

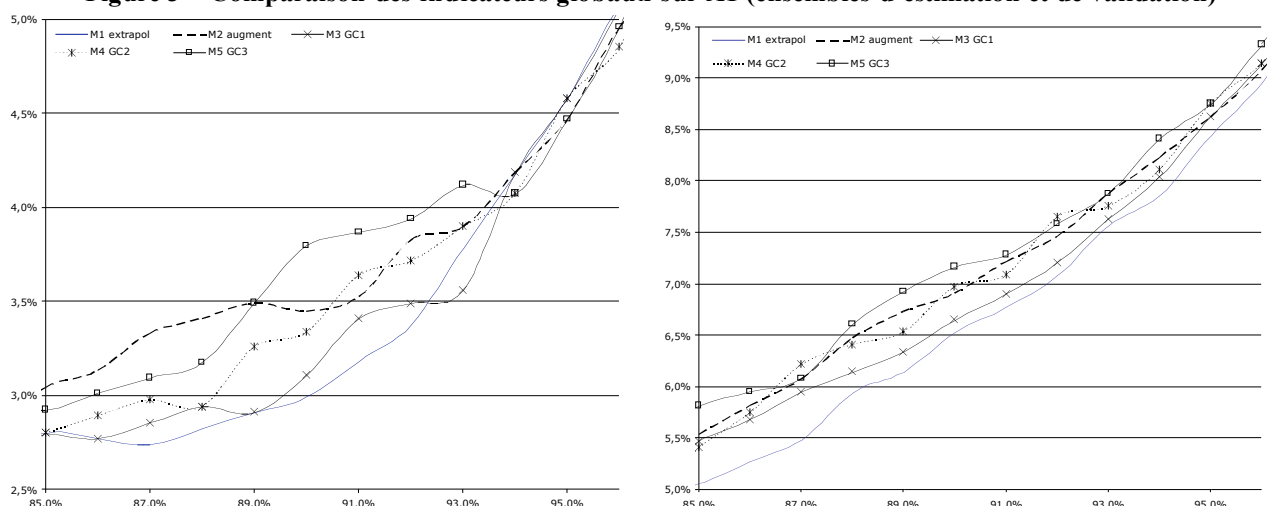

**Figure 4 - Comparaison des taux de défaut (sur A1 à gauche et A2 à droite)**

Enfin, nous appliquons le "meilleur" modèle choisi sur A1 à la population test A2 et mesurons les performances obtenues (fig. 4, à droite). Les performances obtenues ne diffèrent pas ici de manière significative (les tailles d'échantillon ne sont pas assez grandes). Cependant, on voit que le modèle M3 correspondant au tirage aléatoire du groupe de contrôle GC1 a un comportement similaire à celui du modèle M1 (extrapolation). Les autres modèles (augmentation ou groupes de contrôle biaisés) ont des comportements similaires entre eux avec des taux d'erreur supérieurs (à la significativité près des résultats). Le traitement des refusés n'apporte encore pas grand'chose ici.

Toutefois, la démarche que nous présentons permet de mettre en oeuvre systématiquement une approche de traitement des refusés, et de comparer les performances des modèles obtenus.

## Conclusion

Nous avons présenté une méthodologie de traitement des refusés ainsi que quelques modèles utilisés classiquement pour le traitement des refusés. En l'absence de méthode théoriquement fondée, on doit construire et évaluer ces techniques conditionnellement aux données disponibles. Les simulations effectuées permettent de comparer ces modèles dans le cadre d'une expérience plausible, et montrent qu'aucun d'entre eux n'est significativement supérieur aux autres, conclusion suggérée par d'autres auteurs (Hand, 2005). Pour comparer les modèles, il est important de pouvoir générer facilement de nombreux modèles statistiques et de choisir un critère de comparaison adapté, prenant en compte le point de fonctionnement : nous avons choisi ici d'utiliser le taux de défaut des acceptés, bien adapté au contexte de l'octroi de crédit.

Lorsque c'est possible, nous préconisons l'utilisation d'un groupe de contrôle soigneusement choisi. Les simulations présentées dans cet article montrent que la méthode de choix du groupe de contrôle a évidemment un impact sur le résultat.

Nous étudions actuellement comment appliquer à la problématique du traitement des refusés les développements récents de la théorie de l'apprentissage statistique, tels que les approches « transductives » et l'apprentissage semi-supervisé. Des travaux complémentaires seraient sans doute nécessaires pour affiner les calculs de barres d'erreur sur les indicateurs de performance retenus.